\pdfoutput=1

\documentclass[11pt]{article}

\usepackage[]{acl}

\usepackage{times}
\usepackage{latexsym}
\usepackage{graphicx} 

\usepackage[T1]{fontenc}

\usepackage[utf8]{inputenc}

\usepackage{microtype}

\usepackage{makecell}

\newcommand{\twostage}{\textsc{A$^2$CU}}
\newcommand{\onestage}{\textsc{A$^3$CU}}

\usepackage{times}
\usepackage{latexsym}

 \usepackage{amssymb}
 \usepackage{amsmath}
 \usepackage{subcaption}
 \usepackage{booktabs}

\title{Towards Interpretable and Efficient Automatic Reference-Based Summarization Evaluation}

\author{
 Yixin Liu $^{1}$ 
 \quad \textbf{Alexander R. Fabbri} $^{2}$
 \quad \textbf{Yilun Zhao}$^{1}$
 \quad \textbf{Pengfei Liu}$^{3}$ 
 \\
 \quad \textbf{Shafiq Joty}$^{2}$ 
 \quad \textbf{Chien-Sheng Wu}$^{2}$ 
 \quad \textbf{Caiming Xiong}$^{2}$
 \quad \textbf{Dragomir Radev}$^{1}$ \\
  $^1$Yale University, 
  $^2$Salesforce AI Research,
  $^3$Shanghai Jiao Tong University  \\
  \texttt{\{yixin.liu,  dragomir.radev\}@yale.edu}
 }

\begin{document}
\maketitle

\begin{abstract}
    Interpretability and efficiency are two important considerations for    
    the adoption of neural automatic metrics.
    In this work, we develop strong-performing automatic metrics for reference-based summarization evaluation, based on a two-stage evaluation pipeline that first extracts basic information units from one text sequence and then checks the extracted units in another sequence.
    The metrics we developed include two-stage metrics that can provide high interpretability at both the fine-grained unit level and summary level, and one-stage metrics that achieve a balance between efficiency and interpretability.
    We make the developed tools publicly available at  \url{https://github.com/Yale-LILY/AutoACU}.
\end{abstract}

\section{Introduction}

Automatic evaluation is an integral part of scaling natural language generation (NLG) system development and evaluation. 
While neural models have seen great success in NLG systems, their adoption in automatic metric development has been much slower, and classic metrics such as ROUGE~\cite{lin-2004-rouge} and BLEU~\cite{papineni-etal-2002-bleu} are still used more often than neural ones~\cite{sellam-etal-2020-bleurt, Zhang*2020BERTScore:, yuan2021bartscore}.
Compared to neural systems, we argue that there are unique requirements for neural metrics to be adopted:
(1) \textbf{\textit{interpretability}} -- the metric scores should be interpretable and provide intuitive insights into system performance and system output quality.
(2) \textbf{\textit{evaluation efficiency}} -- ideally, the automatic metrics should only introduce a small computation overhead, since it should be possible to use them on the fly for system development and fine-tuning.
In this work, we aim to design neural metrics that are more aligned with these requirements, which we believe will facilitate adoption.

Our method focuses on a two-step decomposition of text sequence comparison following related work~\cite{bhandari-etal-2020-evaluating, zhang-bansal-2021-finding, Liu2022RevisitingTG} --  
first dissecting the information in one text sequence into multiple simple facts and then checking the presence of these facts in another text sequence.
Specifically, our automatic evaluation pipeline mirrors the human evaluation protocol of \citet{Liu2022RevisitingTG}, which uses atomic content units, or ACUs, as the simple facts for comparing text sequences.
We believe such a \textbf{two-stage automatic evaluation} can be more \textbf{\textit{interpretable}} and \textit{\textbf{transparent}}.
Specifically, at the ACU level, the evaluation result indicates the presence or absence of an extracted information unit; at the summary level, the aggregated ACU score represents the percentage of information overlap from one text sequence to another.
In contrast, it can be difficult to interpret the results of certain neural metrics such as BERTScore~\cite{Zhang*2020BERTScore:} or BARTScore~\cite{yuan2021bartscore}.
For example, BARTScore assigns the normalized log-likelihood as the text similarity score, which is non-positive and non-linear, making it difficult to understand the system output quality based on its metric score.

Despite their advantages, such two-stage metrics~\cite{deutsch-etal-2021-towards, zhang-bansal-2021-finding, fabbri-etal-2022-qafacteval} can be much slower to run, even slower than the evaluated systems. 
Therefore, apart from the two-stage evaluation method, we also propose a more \textbf{\textit{efficient}} one-stage metric that directly predicts the aggregated summary-level scores by training on the recently proposed RoSE~\cite{Liu2022RevisitingTG} benchmark.
The one-stage metric retains the summary-level interpretability as in the two-stage evaluation, striking a balance between efficiency and interpretability.
We also explore using the two-stage evaluation as pre-training for the one-stage metric, which further improves performance.

Our contributions can be summarised as:
(1) A fine-grained two-stage automatic metric for reference-based summarization evaluation, which provides high interpretability;
(2) An efficient one-stage automatic metric, which offers a balance between interpretability and evaluation efficiency;
(3) Both types of metrics we developed achieve state-of-the-art performance on summarization evaluation, and we make them publicly available and release them as an easy-to-use Python package.\footnote{We provide the metric training scripts and a Python package for trained metrics at \url{https://github.com/Yale-LILY/AutoACU}.}

\section{Preliminaries}
\paragraph{Information Similarity for Summarization Evaluation}
The most common and important evaluation method of summarization systems is assessing the \textit{similarity} between system-generated summaries and reference summaries.
Since there are no widely accepted definitions of such similarity among the related work~\cite{nenkova-passonneau-2004-evaluating, lin-2004-rouge, bhandari-etal-2020-evaluating, 10.1162/tacl_a_00373, zhang-bansal-2021-finding}, we refer it as the \textbf{\textit{information similarity}} in this work -- two completely similar text sequences should convey exactly the same information to the users, 
following the suggestion of \citet{deutsch-roth-2021-understanding}.

\paragraph{Automatic Metrics of Information Similarity}
Traditional automatic metrics of information similarity compare the lexical overlap of two text sequences, such as ROUGE~\cite{lin-2004-rouge}, BLEU~\cite{papineni-etal-2002-bleu}.
In contrast, a family of neural automatic metrics (e.g., BERTScore~\cite{Zhang*2020BERTScore:}, MoverScore~\cite{zhao-etal-2019-moverscore}, BARTScore~\cite{yuan2021bartscore}) focus more on semantic similarity by leveraging the pre-trained language models.
Neural metrics can also be pre-trained on pseudo training signals~\cite{sellam-etal-2020-bleurt, zhong-etal-2022-towards} or related corpus~\cite{yuan2021bartscore, gao-etal-2021-simcse}, or be supervisedly fine-tuned~\cite{rei-etal-2020-comet}.
Apart from the one-stage metrics, related work on two-stage metrics proposes methods of decomposing the evaluation process into finer-grained sub-tasks, such as the QA-based QAEval~\cite{deutsch-etal-2021-towards} metric, and the Lite$^3$Pyramid~\cite{zhang-bansal-2021-finding} metric, which automates the evaluation process of LitePyramid~\cite{shapira-etal-2019-crowdsourcing} protocol.

\paragraph{Evaluation of Automatic Metrics}
To evaluate information similarity metrics for text summarization, a few human evaluation benchmarks~\cite{bhandari-etal-2020-evaluating, 10.1162/tacl_a_00373, zhang-bansal-2021-finding, Liu2022RevisitingTG} have been collected, which contain system-generated summaries and their human evaluation scores.  
Automatic metric performance is measured by the correlation between the automatic metric scores and human evaluation scores of the system-generated summaries.
For text summarization metrics, such correlations can be calculated at the \textit{system} level and the \textit{summary} level.
More specifically, given $n$ input articles and $m$ summarization systems, the human evaluation and an automatic metric result in two $n$-row, $m$-column score matrices $H$, $M$ respectively.
The \textit{summary}-level correlation is an average of sample-wise correlations:
\begin{equation}
\label{eq:summ_corr}
    r_{\mathrm{sum}}(H, M) = \frac{\sum_i \mathcal{C}(H_i, M_i)}{n}, 
\end{equation}
where $H_i$, $M_i$ are the evaluation results on the $i$-th data sample and $\mathcal{C}$ is a function calculating a correlation coefficient (e.g., the Pearson's correlation coefficient).
In contrast, the \textit{system}-level correlation is calculated on the aggregated system scores:
\begin{equation}
\label{eq:sys_corr}
    r_{\mathrm{sys}}(H, M) = \mathcal{C}(\bar{H}, \bar{M}), 
\end{equation}
where $\bar{H}$ and $\bar{M}$ contain $m$ entries which are the average system scores across $n$ data samples, e.g., $\bar{H}_0 = \sum_i H_{i,0} / n$.

\section{Methods}

We first describe our approach of a two-stage decomposition of automatic information similarity evaluation -- (1) extracting fine-grained content units from one text sequence; (2) checking the existence of the extracted units in another text sequence. 
Then we introduce methods of training a one-stage automatic metric for information similarity and using the two-stage decomposition for pre-training. 

\subsection{Two-Stage Evaluation}

\begin{figure}[t!]
    \centering
    \includegraphics[width=1.0\linewidth]{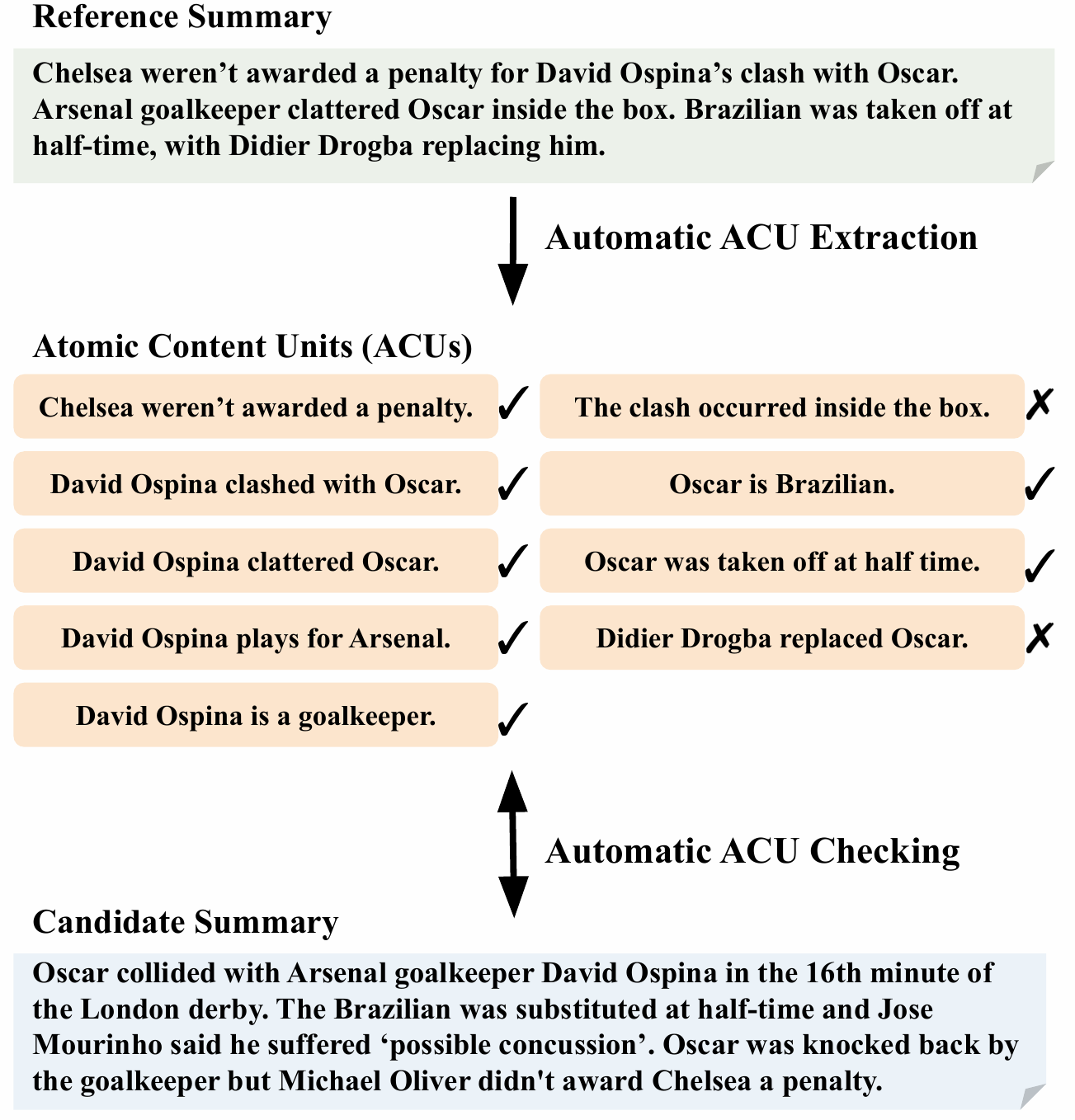}
 \caption{\label{fig:example}Example of two-stage automatic summarization evaluation based on the Atomic Content Unit (ACU) protocol.
 In the first stage, an automatic ACU extraction model dissects the information in one text sequence into ACUs.
 In the second stage, an automatic ACU checking (matching) model checks the presence of the extracted ACUs against another text sequence.
 We use the example provided by \citet{Liu2022RevisitingTG}.
    }
\end{figure}

In Fig.~\ref{fig:example}, we provide an example of the two-stage automatic summarization evaluation process of calculating the \textit{recall} score of a system-generated summary using the reference summary.

\noindent \textbf{Content Unit Extraction} A (long) text sequence can contain more than one fact, or simple information unit.
Therefore, we follow \citet{Liu2022RevisitingTG} by using \textit{Atomic Content Units} (ACUs) to refer to the basic information units. 
We formulate automatic ACU extraction as a sequence-to-sequence (Seq2Seq) problem~\cite{10.5555/2969033.2969173}:
$A \leftarrow g(S)$,
where $S$ is the input text sequence, $A$ is a concatenation of a set of ACUs $\mathcal{A}$ generated as a sequence, and $g$ is a Seq2Seq model.

\noindent \textbf{Content Unit Checking}
Having extracted a set of ACUs $\tilde{\mathcal{A}}$ from one text sequence $S_1$, we use a Natural Langauge Inference (NLI)~\cite{gururangan-etal-2018-annotation} model $f$ to check if the information in an extracted ACU $a$ is conveyed by another text sequence $S_2$:
\begin{equation}
\label{eq:nli-two}
\small
    l_a = f(S_2, a) = \begin{cases}
    1 & \textrm{if $a$ is entailed by $S_2$}, \\
    0 & \textrm{otherwise},
    \end{cases}
\end{equation}
where $l_a$ is the label of $a$ assigned by the model $f$.
In addition to this standard NLI setting of viewing $S_2$ as the premise and $a$ as the hypothesis, we also explore adding $S_1$ as part of the model input serving as context: 
\begin{equation}
\label{eq:nli-three}
\small
    l_a = f^*(S_2, a | S_1) = \begin{cases}
    1 & \textrm{if $a$ is entailed by $S_2$}, \\
    0 & \textrm{otherwise},
    \end{cases}
\end{equation}

We use BERT~\cite{devlin-etal-2019-bert} as the NLI model architecture and follow its input format for the standard setting (Eq.~\ref{eq:nli-two}), which is a concatenation of $S_2$ and $a$.
For the extended setting (Eq.~\ref{eq:nli-three}), we define the input as a concatenation of $S_1$, $S_2$, $a$.

Based on these two stages, we can define a \textit{recall} information similarity score of $S_2$ w.r.t. $S_1$:
\begin{equation}
\label{eq:two-stage-recall}
\small
    \mathcal{R}(S_2|S_1) = \frac{\sum_{a \in \tilde{\mathcal{A}}} l_a}{|\tilde{\mathcal{A}}|}.
\end{equation}

\subsection{One-Stage Metric and Its Pre-Training}

To improve evaluation efficiency, we propose using our two-stage approach to generate pre-training data for a more lightweight, one-stage metric. 
Specifically, we use BERT as the backbone for a scoring model $h$ to approximate the two-stage recall score (Eq.~\ref{eq:two-stage-recall}):
\begin{equation}
\small
\label{eq:one-stage-recall}
    \mathcal{R}_h(S_2|S_1) \leftarrow \mathcal{R}(S_2|S_1).
\end{equation}
The model is trained with the mean squared error between the target 
score and the predicted score.
The input format is ``[CLS]$S_2$[SEP]$S_1$[SEP]'' following \citet{devlin-etal-2019-bert}, and a single linear layer is introduced to map the hidden representation of ``[SEP]'' into the predicted numeral score.
We can also define an F1 score as 
\begin{equation}
\label{eq:one-stage-f1}
\small
    \mathcal{F}_h(S_2|S_1) = \frac{2 \mathcal{R}_h(S_2|S_1)  \mathcal{R}_h(S_1|S_2)}{ \mathcal{R}_h(S_2|S_1) +  \mathcal{R}_h(S_1|S_2) }.
\end{equation}

\noindent \textbf{Pre-training Corpora} 
As shown by \citet{sellam-etal-2020-bleurt}, the robustness of automatic metrics can be improved by pre-training with synthetic pre-training data.
We further extend this approach following the finding of related work~\cite{liu-liu-2021-simcls, liu-etal-2022-brio}, which shows that pre-trained summarization models such as BART can generate diverse and high-quality candidate summaries and summarization models can benefit from contrastive learning with the generated candidates. 
In a similar spirit, we construct the pre-training corpora on the existing summarization datasets such as CNN/DailyMail~\cite{Nallapati:16}, and for each data example we generate multiple candidate summaries using a fine-tuned summarization model.
The generated summaries are then scored by the two-stage evaluation method (Eq.~\ref{eq:two-stage-recall}), which are used for the pre-training of one-stage model $h$.

\section{Experiments}

\subsection{Experimental Settings}

\paragraph{Datasets} We mainly use a recently-introduced summarization evaluation benchmark, RoSE~\cite{Liu2022RevisitingTG}, for automatic metric development and evaluation.
It contains human evaluation \textit{recall} scores of the system-generated summaries based on the reference summaries w.r.t. the information similarity on three summarization datasets, CNN/DailyMail (CNNDM)~\cite{Nallapati:16}, XSum~\cite{narayan-etal-2018-dont}, and SamSum~\cite{gliwa-etal-2019-samsum}.
The human evaluation is conducted following the ACU evaluation protocol~\cite{Liu2022RevisitingTG}, and the dataset also provides human-written ACUs on the reference summaries and the associated results (ACU labels) of checking the ACUs against system-generated summaries.  
The statistics are in Tab.~\ref{tab:data-statistics}.

\begin{table}[t!]
\small
\centering
\addtolength{\tabcolsep}{-1pt} 
\begin{tabular}{lccccc}
\toprule
\textbf{Dataset} & \textbf{Split} & \textbf{\#Doc.} & \textbf{\#Sys.}  & \textbf{\#ACU} & \textbf{\#Summ.}\\
\midrule
CNNDM & Test & 500 & 12 & 5.6k & 6k \\
CNNDM & Valid & 1,000 & 8 & 11.6k & 8k \\
XSum & Test & 500 & 8 & 2.3k & 4k \\
SamSum & Test & 500 & 8 & 2.3k & 4k\\
\bottomrule
\end{tabular}
\addtolength{\tabcolsep}{1pt} 
\caption{Statistics of the RoSE dataset as reported in \citet{Liu2022RevisitingTG}.
\textbf{\#ACU} and \textbf{\#Summ.} are the number of annotations at ACU and summary levels.
}
\label{tab:data-statistics} 
\end{table}

\paragraph{Baseline Metrics}
We compare our methods with related automatic metrics for text sequence comparison.
Since the RoSE benchmark provides \textit{recall} information similarity scores, only metrics with recall scores are compared.

\noindent \textbf{ROUGE}~\cite{lin-2004-rouge} compares two text sequences by the $n$-grams overlap between them.
We report the performance of its ROUGE-1/2/L variants.

\noindent \textbf{BERTScore}~\cite{Zhang*2020BERTScore:} measures text sequence similarity 
using hidden representations computed by pre-trained language models such as BERT~\cite{devlin-etal-2019-bert}.
We report its variants with RoBERTa~\cite{Liu2019RoBERTaAR} (BERTScore$_R$) and DeBERTa~\cite{he2021deberta}  (BERTScore$_D$).

\noindent \textbf{BARTScore}~\cite{yuan2021bartscore} interprets the similarity of text sequence $x$ to $y$ as the probability of $x$ given $y$ predicted by a pre-trained language model such as BART~\cite{lewis-etal-2020-bart}.
We report its variants pre-trained on CNNDM (BARTScore$_C$) and ParaBank2~\cite{hu-etal-2019-large} (BARTScore$_P$).

\noindent \textbf{QAEval}~\cite{deutsch-etal-2021-towards} measures information similarity by question answering accuracy.
We report both its exact match score (QAEval$_{EM}$) and F1 score (QAEval$_{F1}$).

\noindent \textbf{Lite$^3$Pyramid}~\cite{zhang-bansal-2021-finding} introduces a similar approach as our two-stage evaluation but uses a semantic role labeling~\cite{he-etal-2017-deep} model to extract content units.
We report its variants that use the predicted label (Lite$^3$Pyramid$_L$) and probability (Lite$^3$Pyramid$_P$) of the two-class NLI model.

\begin{table}[t!]
\small
\centering
\addtolength{\tabcolsep}{-2pt} 
\begin{tabular}{@{\extracolsep{1pt}}lcccccc@{}}
\toprule
 & \multicolumn{2}{c}{\textbf{CNNDM}} & \multicolumn{2}{c}{\textbf{XSum}} & \multicolumn{2}{c}{\textbf{SamSum}} \\
 &  \textbf{Sys.} & \textbf{Sum.} & \textbf{Sys.} & \textbf{Sum.} & \textbf{Sys.} & \textbf{Sum.} \\
 \cmidrule{2-3} \cmidrule{4-5} \cmidrule{6-7} 
ROUGE1                   & .788        & .468         & .714       & .293        & .929         & .439          \\
 ROUGE2                   & .758        & .453         & .643       & .266        & \textbf{1.00}        & .395          \\
 ROUGEL                   & .879        & .454         & .643       & .258        & .929         & .415          \\
 BERTScore$_R$       & .515        & .448         & .571       & .277        & .857         & .417          \\
 BERTScore$_D$       & .424        & .424         & .571       & .262        & .857         & .409          \\
 BARTScore$_C$    & .727        & .435         & .643       & .260        & .929         & .438          \\
 BARTScore$_P$ & .727        & .453         & .714       & .282        & .929         & .430          \\
  QAEval$_{EM}$                & .515        & .296         & .357       & .149        & .857         & .352          \\
 QAEval$_{F1}$                 & .849        & .358         & .429       & .198        & .929         & .384          \\
 Lite$^3$Pyramid$_L$                       & .849        & .466         & .643       & .207        & \textbf{1.00}        & .494          \\
 Lite$^3$Pyramid$_P$                      & .849        & .452         & .714       & .245        & \textbf{1.00}        & .467          \\
\midrule
\twostage$_{P}$       & .879        & .521$^\dag$         & \textbf{.786}       & .336$^\dag$         & \textbf{1.00}        & .532$^\dag$           \\
 \twostage$_{F}$              & .818        & .557$^\dag$         & \textbf{.786}       & \textbf{.348}$^\dag$         & \textbf{1.00}        & \textbf{.556}$^\dag$           \\
 \twostage$_{FC}$                   & .879        & .555$^\dag$         & .214       & .202        & .571         & .275          \\
 \onestage$_F$           & \textbf{.909}        & .493$^\dag$         & \textbf{.786}       & .299        & .929         & .445          \\
 \onestage$_P$             & .879        & .558$^\dag$         & \textbf{.786}       & .307        & .929         & .472          \\
 \onestage$_{PF}$   & .879        & \textbf{.564}$^\dag$         & \textbf{.786}       & .319$^\dag$         & .929         & .474          \\
 \bottomrule
\end{tabular}
\addtolength{\tabcolsep}{2pt} 
\caption{The Kendall's correlation between the automatic metric scores and human evaluation scores on the RoSE dataset.
The correlation is calculated at both the system level and the summary level. 
We use the \textit{recall} score of the automatic metrics.
\twostage~is our two-stage evaluation method, \onestage~is our one-stage metric.
$\dag$: significantly ($p < 0.05$) better than the best baseline. 
The baseline results are from \citet{Liu2022RevisitingTG}.
}
\label{tab:metric-corr} 
\end{table}

\paragraph{Implementation Details}
For the \textit{two-stage} evaluation, the ACU extraction model is based on a T0~\cite{sanh2022multitask}
model\footnote{\url{huggingface.co/bigscience/T0_3B}} fine-tuned on the human-written ACUs provided in RoSE.
As for the NLI model for ACU checking (Eq.~\ref{eq:nli-two}\&\ref{eq:nli-three}), we use a pre-trained DeBERTa~\cite{he2021deberta} NLI model\footnote{\url{https://huggingface.co/microsoft/deberta-xlarge-mnli}.} as the start point and further fine-tune it on the RoSE dataset with the available gold standard ACU labels.
We name the two-stage method \twostage~(\textbf{A}uto\textbf{ACU}), and it has three variants in total: \twostage$_{P}$ is with the pre-trained NLI model, \twostage$_{F}$ is with the fine-tuned NLI model (Eq.~\ref{eq:nli-two}), \twostage$_{FC}$ is with the fine-tuned NLI model taking the source text as part of the input.

For the pre-training of \textit{one-stage} metric, we use system-generated summaries generated by a pre-trained BART model on the CNNDM dataset.
For each data example, 12 summaries are scored by the two-stage evaluation method and used to pre-train the one-stage metric. 
After the pre-training, the one-stage metric can be further fine-tuned on the gold-standard scores.
We name the one-stage metric as \onestage~(\textbf{A}ccelerated\textbf{A}uto\textbf{ACU}), and it has three variants: \onestage$_F$ is directly fine-tuned on RoSE, \onestage$_P$ is pre-trained with \twostage~only, \onestage$_{PF}$ is first pre-trained then fine-tuned.
We note that the training on the RoSE dataset is performed on the validation split of CNNDM.
Regarding \textit{\textbf{metric efficiency}}, \onestage~takes only 6\% the inference time of \twostage, demonstrating its superior efficiency.
More details are in Appendix \ref{appendix:details}.

\begin{table}[t!]
\small
\centering
\addtolength{\tabcolsep}{-2pt} 
\begin{tabular}{@{\extracolsep{1pt}}lcccccc@{}}
\toprule
 & \multicolumn{2}{c}{\textbf{CNNDM}} & \multicolumn{2}{c}{\textbf{XSum}} & \multicolumn{2}{c}{\textbf{SamSum}} \\
 &  \textbf{Sys.} & \textbf{Sum.} & \textbf{Sys.} & \textbf{Sum.} & \textbf{Sys.} & \textbf{Sum.} \\
 \cmidrule{2-3} \cmidrule{4-5} \cmidrule{6-7} 
Semi\twostage    & .849        & .647         & .786       & .450        & 1.00        & .621          \\
 \twostage$_{F}$              & .818        & .557         & .786     & .348         & 1.00     & .556           \\
 \bottomrule
\end{tabular}
\addtolength{\tabcolsep}{2pt} 
\vspace{-2mm}
\caption{Performance comparison between \twostage$_{F}$ and the semi-automatic metric (Semi\twostage) using reference ACUs.
Kendall's correlation coefficients are reported.
}
\vspace{-3mm}
\label{tab:compare} 
\end{table}

\subsection{Results}
\label{subsec:4-2}

We report the results in Tab.~\ref{tab:metric-corr}.
Kendall's correlation coefficients are used to evaluate the metric performance at both the system and summary levels, which shows the following:
(1) Both our two-stage and one-stage metrics can outperform the baseline methods across three datasets.
(2) The improvement of our metrics is more significant at the summary level than at the system level.
(3) Compared with the one-stage metrics, our two-stage metrics generalize better on XSum and SamSum datasets.
(4) The pre-training is effective for the one-stage metric to achieve strong performance since \onestage$_P$ can outperform \onestage$_F$ at the summary level without fine-tuning on the RoSE dataset.

\subsection{Analysis}

\begin{table}[t!]
\small
\centering
\begin{tabular}{lccc}
\toprule
 \textbf{Dataset}   &   \textbf{Precision} &   \textbf{Recall} &    \textbf{F1} \\
\midrule
 CNNDM    &       83.63 &    79.84 & 81.69 \\
 XSum    &       80.11 &    79.14 & 79.62 \\
 SamSUM    &    87.01 &    87.14 & 87.08 \\ 
 \bottomrule
\end{tabular}
\vspace{-2mm}
\caption{Quality analysis of \twostage-generated ACUs compared with reference ACUs using ROUGE-1 scores.
}
\vspace{-3mm}
\label{tab:acu-quality} 
\end{table}

\paragraph{Performance Analysis of \twostage}
We compare \twostage~performance with Semi\twostage, a \textit{semi}-automatic metric that follows the same scoring mechanism but uses the reference ACUs as the NLI model input (Eq.~\ref{eq:nli-two}).
Results in Tab.~\ref{tab:compare} show that using the reference ACUs yields better performance, suggesting that better generated ACU quality may lead to further improvement.
To analyze this quality, we calculate its similarity with reference ACUs using ROUGE scores between them in a greedy-matching manner. 
Specifically, given a set of generated ACUs $\tilde{\mathcal{A}}$ and reference ACUs $\mathcal{A}$, we calculate the example-level \textit{recall} score as:
\begin{equation}
\small
    r(\tilde{\mathcal{A}}|\mathcal{A}) = \frac{\sum_{\tilde{a} \in \tilde{\mathcal{A}}} \max_{a \in \mathcal{A}}\mathrm{ROUGE}(a, \tilde{a})}{|\tilde{\mathcal{A}}|}.
\end{equation}
Similarly, we can define the precision score and F1 score.
The results on three dataset splits are reported in Tab.~\ref{tab:acu-quality}, showing a high degree of similarity between the reference and generated ACUs.

\noindent \textbf{Comparative Study of \onestage}
\begin{table}[t!]
\small
\centering
\addtolength{\tabcolsep}{-2pt} 
\begin{tabular}{@{\extracolsep{1pt}}lcccccc@{}}
\toprule
 & \multicolumn{2}{c}{\textbf{CNNDM}} & \multicolumn{2}{c}{\textbf{XSum}} & \multicolumn{2}{c}{\textbf{SamSum}} \\
 &  \textbf{Sys.} & \textbf{Sum.} & \textbf{Sys.} & \textbf{Sum.} & \textbf{Sys.} & \textbf{Sum.} \\
 \cmidrule{2-3} \cmidrule{4-5} \cmidrule{6-7} 
\onestage-B    & .879        & .558         & .786       & .315        & .929         & .467          \\
 \onestage-R  & .879        & .483         & .643       & .290        & .929         & .429          \\
 \onestage   & .879        & .564         & .786       & .319        & .929         & .474          \\
 \bottomrule
\end{tabular}
\addtolength{\tabcolsep}{2pt} 
\caption{Performance comparison of different variants of the \onestage~metric fine-tuned on RoSE.
\onestage~is based on BERT large model and pre-trained on the two-stage evaluation method (\twostage).
\onestage-B is its counterpart based on BERT base model, 
\onestage-R is the counterpart pre-trained to predict ROUGE scores.
The baseline results are from \citet{Liu2022RevisitingTG}.
}
\vspace{-3mm}
\label{tab:analysis} 
\end{table}
We investigate the impact of two design decisions for the one-stage metric, the model size, and the supervision signal used in pre-training. 
To this end, we show metric performance with two model sizes and the metric performance when it is pre-trained to predict the average score of ROUGE-1/2/L.
The results in Tab.~\ref{tab:analysis} suggest that larger model sizes and more suitable pre-training can improve metric performance.

\noindent \textbf{Evaluation on Other Related Benchmarks}
We provide additional evaluation results of our F1-based automatic metrics (e.g., Eq.~\ref{eq:one-stage-f1}) in Appendix~\ref{appendix:compare}.
Apart from RoSE, we evaluate the metrics on two related benchmarks, the STS (Semantic Textual Similarity) benchmark~\cite{cer-etal-2017-semeval} and WMT19 Metrics Shared Task \textsc{daRR} benchmark~\cite{ma-etal-2019-results}, which provides a more comprehensive investigation of the generalization ability of our proposed metrics.

\section{Conclusions}

We develop high-performing reference-based summarization automatic metrics, including two-stage metrics providing fine-grained interpretability and one-stage metrics for a balance between efficiency and interpretability.
Furthermore, we show that the two-stage metric can be used to effectively pre-train the one-stage metric, helping to mitigate the data scarcity in developing automatic metrics.

\section*{Acknowledgements}
We are grateful to the anonymous reviewers for their constructive comments. 
We thank Arman Cohan for insightful discussions and suggestions.

\section*{Limitations}

Our metrics are developed and evaluated on English corpora only, and it remains unclear whether the metrics can achieve consistent performance in other languages.
The two-stage metrics we proposed are based on relatively large language models and it can be time-consuming and computationally expensive to use them, especially on large corpora.

\bibliography{anthology,custom}
\bibliographystyle{acl_natbib}
\appendix

\section{Experiment Details}
\label{appendix:details}
For the two-stage evaluation, the ACU extraction model is a T0 model containing 3 billion parameters, and the ACU checking model is a pre-trained DeBERTa model with 750 million parameters.
Our one-stage metric is based on BERT, and its two variants have 340 and 110 million parameters respectively. 
We conduct experiments on NVIDIA RTX A6000 GPUs, and the experiments take around 30 GPU hours to finish.

\paragraph{Inference Efficiency}
In Tab.~\ref{tab:speed} we compare the inference time of \onestage~and \twostage.
\onestage~only takes around 6\% of the inference time of \twostage.

\begin{table}[t]
\small
\centering
\begin{tabular}{lccc}
\toprule
 \textbf{Dataset}   &   \textbf{\twostage-Stage1} &  \textbf{\twostage-Stage2} &   \textbf{\onestage} \\
\midrule
 CNNDM    &       21.3 &    14.7 & 2.34 \\
 XSum    &       9.4 &    1.9 & 0.56 \\
 SamSUM    &    9.3 &    1.9 & 0.69 \\ 
 \bottomrule
\end{tabular}
\caption{Inference time (in minutes) comparison of \onestage~and \twostage. For \twostage~we show the time of ACU generation (Stage 1) and matching (Stage 2).
}
\label{tab:speed} 
\end{table}

\section{Evaluation on Related Information Similarity Benchmark}
\label{appendix:compare}

Apart from the RoSE benchmark, we evaluate metric performances on two related benchmarks for assessing the \textit{information similarity} between two text sequences:
the STS (Semantic Textual Similarity) benchmark~\cite{cer-etal-2017-semeval} and the human evaluation results from the WMT19 Metrics Shared Task~\cite{ma-etal-2019-results}.

\paragraph{STS Benchmark} STS benchmark\footnote{The data is provided at \url{http://ixa2.si.ehu.eus/stswiki/index.php/STSbenchmark}.} contains English sentence pairs and the associated human-annotated semantic similarity scores~\cite{cer-etal-2017-semeval}, which has a similar evaluation target as the information similarity for text summarization evaluation. 
With this benchmark, the semantic similarity metrics are evaluated by the correlation between the predicted similarity scores and the reference scores.
Following the previous work~\cite{gao-etal-2021-simcse, reimers-etal-2016-task}, we report Spearman's correlation coefficients on the test split of the benchmark containing around 1.5k data examples.
Since the metrics we evaluated previously are not designed specifically for the STS task, we compare another metric, SimCSE~\cite{gao-etal-2021-simcse}, which achieves strong performance on this task.

\paragraph{WMT19 \textsc{daRR} Benchmark} 
This benchmark contains human-annotated scores of system-generated translations~\cite{ma-etal-2019-results}. 
We use only the to-English part of the benchmark, which results from \textit{reference-based} direct assessment (DA) of translation quality.
The human-annotated DA scores are then transformed into relative rankings between two translations, i.e., the \textsc{daRR} scores.\footnote{We use the pre-processed data provided by \citet{rei-etal-2020-comet} at \url{https://github.com/Unbabel/COMET/}.}
We follow the evaluation setting of \citet{ma-etal-2019-results} by using Kendall’s
Tau-like correlation to evaluate the \textbf{segement-level} performance of automatic metrics. 
The benchmark contains the translations from seven languages to English and results in around 21k translated sentence pairs.

Apart from these two benchmarks, we also report the metric performance under \textit{normalized} ACU scores in RoSE, which are decorrelated with summary lengths to evaluate F1-based metrics~\cite{Liu2022RevisitingTG}.
We note that unlike in \S\ref{subsec:4-2}, all the metrics compared here are F1-based.
In particular, for \onestage~we follow Eq.~\ref{eq:one-stage-f1} to calculate the F1 scores.
The two-stage metrics including \twostage~are not reported because of their lower efficiency.

\begin{table}[t!]
\small
\centering
\addtolength{\tabcolsep}{-3pt} 
\begin{tabular}{lccccc}
\toprule
 & \textbf{STS} & \textbf{WMT} & \textbf{CNNDM} & \textbf{XSum} & \textbf{SamSum} \\
 \midrule
 ROUGE1                  & .578 & .105 &   .403 &  .278 &    .399 \\
 ROUGE2                 & .452 & .071 &   .375 &  .253 &    .343 \\
 ROUGEL                  & .556 & .113 &   .378 &  .252 &    .365 \\
BERTScore$_R$    & .569 & .254 &   .386 &  .280  &    .401 \\
BERTScore$_D$    & .523 & .259 &   .386 &  .274 &    .401 \\
 BARTScore$_C$     & .607 & .254 &   .296 &  .228 &    .341 \\
  BARTScore$_P$    & .695 & \textbf{.266} &   .342 &  .259 &    .371 \\
 SimCSE                 & \textbf{.827} & .235 &   .300   &  .255 &    .356 \\
 \midrule
 \onestage$_P$         & .731 & .209 &   \textbf{.447} &  .298 &    .418 \\
 \onestage$_{PF}$  & .787 & .220  &   .463 &  \textbf{.316} &    \textbf{.427} \\     
 \bottomrule
\end{tabular}
\addtolength{\tabcolsep}{3pt} 
\caption{Metric performance on related benchmarks.
\textbf{STS} is the STS benchmark with Spearman's correlation coefficients.
\textbf{WMT} is the WMT19 \textsc{daRR} benchmark with Kendall's Tau-like correlations.
\textbf{CNNDM}, \textbf{XSum}, \textbf{SamSum} correspond to Kendall's correlation coefficients based on the normalized ACU scores on the RoSE benchmark on different test splits. 
All the correlations are calculated at the \textit{segment} level. 
We use the F1 score of the automatic metrics.
}
\label{tab:metric-other} 
\end{table}

\begin{figure}[t!]
    \centering
         \includegraphics[width=1.0\linewidth]{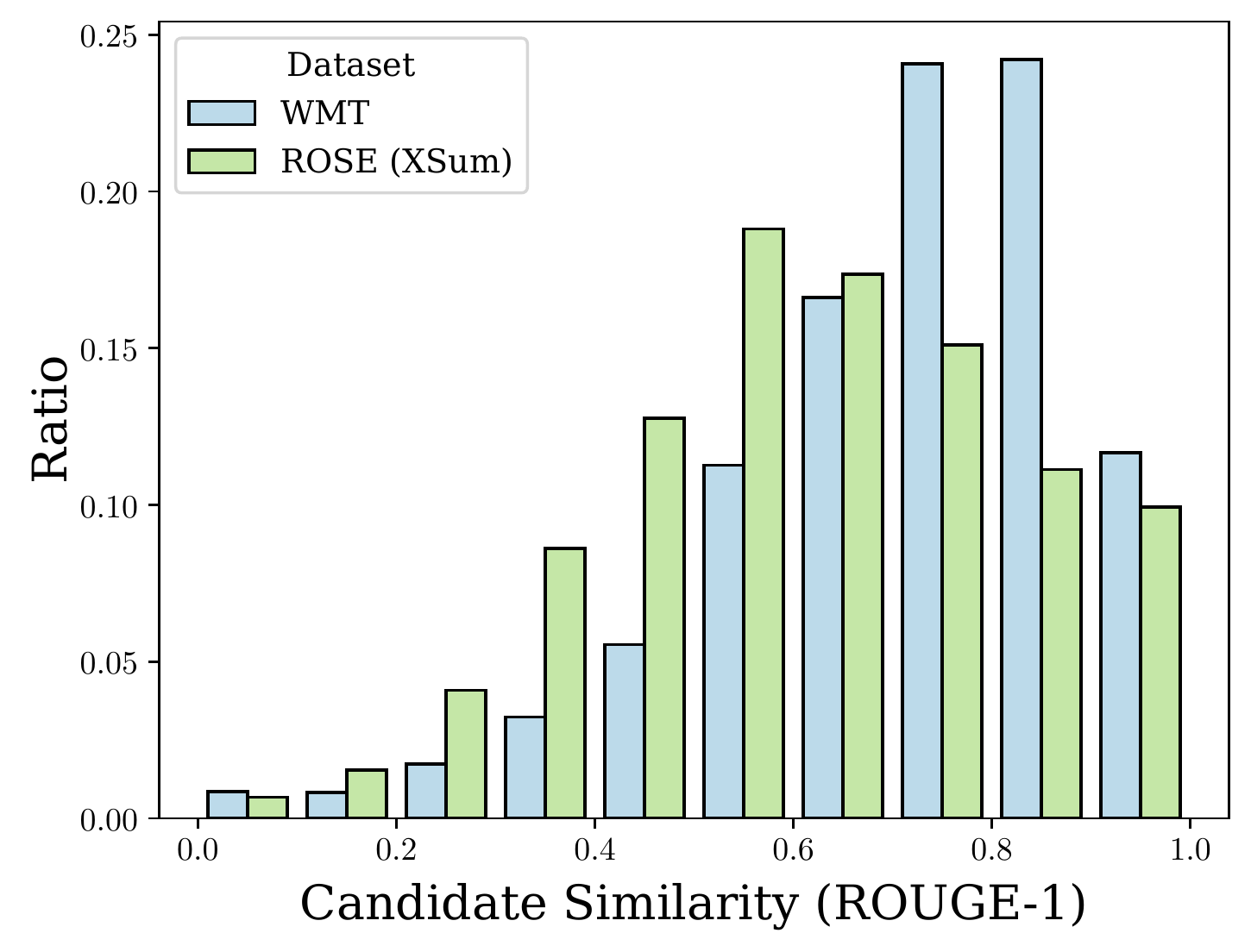}
 \caption{\label{fig:compare}Comparison of candidate outputs similarity in the WMT19 \textsc{daRR} benchmark and the XSum test split of RoSE benchmark.
 The similarity of candidate output pairs of the same example is evaluated by ROUGE1.
    }
\end{figure}

The results in Tab.~\ref{tab:metric-other} show that:

\noindent (1) While the compared metrics have a similar evaluation target, i.e., \textit{information similarity} between text sequences, their performance varies across different benchmarks and there is no single metric that can consistently outperform others.
A similar finding in \citet{ma-etal-2019-results} shows that Pearson's correlation between human-annotated cross-lingual STS scores and machine translation quality estimation scores is only 0.41.

\noindent (2) On the STS benchmark, \onestage~outperforms other metrics except for SimCSE which is specifically designed for the STS task, which we believe results from \onestage's better interpretability thanks to its underlying evaluation process.
Specifically, \onestage's computed scores directly indicate the information overlap between text sequences, which is close to the definition of the STS task.

\noindent (3) On the WMT benchmark, \onestage~fails to outperform BERTScore and BARTScore.
We hypothesize this is because \onestage~is relatively insensitive to the minor differences in candidate translations when implicitly comparing them based on the same reference translation as we found that the system-generated translations in the WMT benchmark have higher similarity than the system-generated summaries in the RoSE benchmark.
Specifically, while both RoSE and WMT19 \textsc{daRR} benchmarks contain reference-based human annotations of candidate output quality, we note that they have different data distributions. 
In detail, the system-generated translations of the same source sentence in the WMT benchmark have higher similarities than the system-generated summaries of the same source articles.
We visualize this discrepancy in Fig.~\ref{fig:compare}, which shows the similarity (as evaluated by ROUGE-1) between different candidate outputs (of the same example) on WMT19 \textsc{daRR} benchmark and the XSum test split\footnote{We chose XSum split because it contains one-sentence news article summaries, which is similar to the data format of the WMT19 benchmark.} of RoSE benchmark.
As illustrated by the figure, candidates in WMT19 benchmark are more similar, which can lead to the performance difference when the same automatic metric is evaluated on these two benchmarks.

\end{document}